\pdfoutput=1
\documentclass{article}
\usepackage[preprint,nonatbib]{neurips_2025}

\usepackage[utf8]{inputenc}
\usepackage[T1]{fontenc}
\usepackage{hyperref}
\usepackage{url}
\usepackage{booktabs}
\usepackage{amsfonts}
\usepackage{nicefrac}
\usepackage{microtype}
\usepackage{xcolor}
\usepackage{graphicx}
\usepackage{amsmath, amssymb, amsthm}
\usepackage{bm}
\usepackage{algorithm}
\usepackage{algpseudocode}
\usepackage{booktabs}
\usepackage{diagbox}
\usepackage{subcaption}
\usepackage{tikz}
\usepackage{graphicx}
\usetikzlibrary{positioning, shapes.geometric, arrows.meta, calc}
\usepackage{multicol, multirow}

\usepackage{amsmath,amsfonts,bm}

\def\eqref#1{equation~\ref{#1}}

\def\1{\bm{1}}

\DeclareMathAlphabet{\mathsfit}{\encodingdefault}{\sfdefault}{m}{sl}
\SetMathAlphabet{\mathsfit}{bold}{\encodingdefault}{\sfdefault}{bx}{n}

\DeclareMathOperator{\sign}{sign}

\newcommand{\norm}[1]{\lVert #1 \rVert}

\title{My Face Is Mine, Not Yours: Facial Protection Against Diffusion Model Face Swapping}

\vspace{-20pt}
\author{%
  Hon Ming Yam \\
  Department of Data Science\\
  City University of Hong Kong\\
  \texttt{hmyam4-c@my.cityu.edu.hk} \\
  \And
  Zhongliang Guo \\
  School of Computer Science\\
  University of St Andrews\\
  \texttt{zg34@st-andrews.ac.uk} \\
  \And
  Chun Pong Lau \\
  Department of Data Science \\
  City University of Hong Kong \\
  \texttt{cplau27@cityu.edu.hk} \\
}

\begin{document}

\vspace{-20pt}
\maketitle
\vspace{-20pt}
\setlength{\abovedisplayskip}{3pt}
\setlength{\belowdisplayskip}{3pt}
\setlength{\abovecaptionskip}{2pt}
\setlength{\belowcaptionskip}{0pt}

\begin{abstract}
    The proliferation of diffusion-based deepfake technologies poses significant risks for unauthorized and unethical facial image manipulation. While traditional countermeasures have primarily focused on passive detection methods, this paper introduces a novel proactive defense strategy through adversarial attacks that preemptively protect facial images from being exploited by diffusion-based deepfake systems. Existing adversarial protection methods predominantly target conventional generative architectures (GANs, AEs, VAEs) and fail to address the unique challenges presented by diffusion models, which have become the predominant framework for high-quality facial deepfakes. Current diffusion-specific adversarial approaches are limited by their reliance on specific model architectures and weights, rendering them ineffective against the diverse landscape of diffusion-based deepfake implementations. Additionally, they typically employ global perturbation strategies that inadequately address the region-specific nature of facial manipulation in deepfakes.
    To overcome these limitations, we propose a novel dual-loss adversarial framework specifically designed for diffusion-based facial deepfakes. Our approach combines a face identity loss that targets the conditional mechanisms underlying diffusion-based face swapping, with an inference-step averaging loss that efficiently addresses the computational challenges inherent in diffusion model optimization. By calculating and averaging deviation losses across multiple inference steps, our method generates unbiased gradient estimations that maximize protective efficacy across varying inference configurations, without requiring prohibitive computational resources. Experimental results demonstrate that our approach provides robust protection against a wide range of diffusion-based deepfake models while maintaining visual fidelity of the protected images, establishing a significant advancement in proactive deepfake countermeasures.
  \end{abstract}
\section{Introduction}\label{sec:Intro}

\begin{figure}[htbp]
    \centering
    \includegraphics[width=\textwidth]{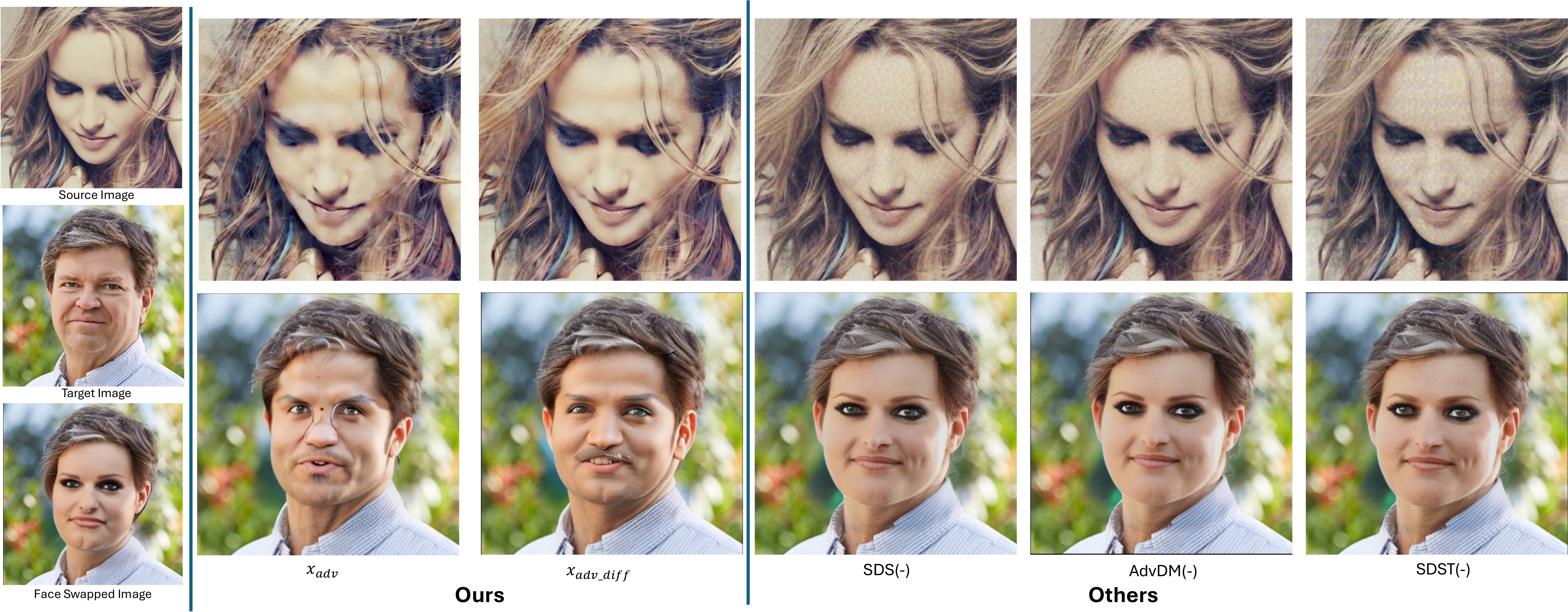}
    \caption{The face swapping result using diffusion-based model pre- and post-protection by our method and other existing methods.}\label{fig:compare}
    \vspace{-20pt}
\end{figure}

The rapid evolution of synthetic media technologies has ushered in an era where digital facial manipulation has become increasingly sophisticated and accessible.
Deepfake technology, particularly face swapping, represents a double-edged sword—offering creative possibilities while simultaneously posing unprecedented threats to privacy, security, and information integrity.
Recent high-profile incidents have demonstrated how deepfakes can be weaponized to spread misinformation, manipulate public discourse, and victimize individuals through non-consensual content creation~\cite{masood2023deepfakes}. As these technologies become more widespread, the need for effective protection mechanisms has never been more urgent.

Recent years have witnessed a paradigm shift in generative modeling, with diffusion models emerging as the dominant framework for high-quality image synthesis~\cite{dhariwal2021diffusion, esser2024scaling, wang2023prolificdreamer, rombach2022high, podell2023sdxl}. Unlike their GAN predecessors, diffusion models generate images through an iterative denoising process that consistently produces results of remarkable fidelity and diversity~\cite{ho2020denoising, song2020score, song2019generative, songDDIM, sohl2015deep}. This technical advancement has rapidly transformed the deepfake landscape, with state-of-the-art face swapping systems increasingly built upon diffusion architectures, as evidenced by recent innovations [DiffFace~\cite{kim2025diffface}, Face Adapter~\cite{han2024faceadapterpretraineddiffusion}, REFace~\cite{baliah2024realistic}, and DiffSwap~\cite{zhao2023diffswap}].
The superior quality and accessibility of these methods have significantly lowered the technical barriers for creating convincing facial manipulations, exacerbating the associated risks.

Despite this shifting landscape, protective countermeasures have failed to keep pace.
Current approaches to combating deepfakes fall into two broad categories: passive detection systems that identify manipulated content after creation, and proactive protection methods that prevent manipulation at the source. While significant progress has been made in detection techniques, their effectiveness remains limited against continuously evolving generative models~\cite{Guillaro_2023_CVPR, tan2024rethinking}. Proactive protection through adversarial perturbations offers a promising alternative by rendering facial images resistant to manipulation. However, existing adversarial protection strategies have predominantly targeted conventional generative frameworks like GANs and autoencoders~\cite{10100731, 10382699, 10447890}, leaving a critical vulnerability against the new generation of diffusion-based deepfakes.

Developing effective adversarial protection against diffusion models presents unique technical challenges that existing approaches fail to address. First, diffusion models employ complex, multi-step generative processes that create computational barriers for traditional adversarial methods, including gradient instability and prohibitive memory requirements~\cite{pmlr-v80-athalye18a, pmlr-v139-yoon21a}. Second, current diffusion-specific adversarial techniques typically require access to specific model architectures and weights, severely limiting their practical applicability against the diverse ecosystem of diffusion-based face swapping implementations. Third, these methods generally apply global perturbation patterns that inadequately address the region-specific nature of facial manipulation in deepfake systems~\cite{pmlr-v202-liang23g, xue2023toward, liang2023mistimprovedadversarialexamples}, where processing pipelines isolate and transform facial regions before reintegration.

To bridge this critical protection gap, we propose a novel adversarial framework specifically designed to defend against diffusion-based face swapping. Our approach introduces two key innovations:
(1) a face identity loss mechanism that directly targets the conditional processing pathways exploited in diffusion-based face manipulation,
and (2) an inference-step averaging technique that efficiently generates robust adversarial perturbations without requiring prohibitive computational resources. By strategically calculating and averaging deviation losses across multiple inference steps, our method produces unbiased gradient estimations that maximize protection efficacy across varying inference configurations, while maintaining the visual quality of protected images.

Through extensive experimental evaluation, we demonstrate that our approach provides robust protection against a diverse range of diffusion-based deepfake systems, including those using different architectures and training methodologies.
Importantly, our method maintains effectiveness even against common adversarial defenses such as JPEG compression, Gaussian blur, and state-of-the-art diffusion-based adversarial purification.
To the best of our knowledge, this work represents the first successful attempt to create a memory-efficient adversarial protection specifically targeting diffusion-based face swapping, addressing a critical vulnerability in current privacy protection frameworks.

Our main contributions can be summarized as follows:
\begin{itemize}
    \item We introduce the first adversarial protection framework specifically designed for diffusion-based face swapping systems, which operates efficiently within restricted GPU memory constraints.
    \item We propose a novel dual-loss optimization strategy that combines face identity preservation with inference-step averaging to overcome the computational challenges inherent in diffusion model optimization.
    \item We demonstrate robust protection against a wide range of diffusion-based face swapping methods, maintaining effectiveness even against common adversarial defenses including JPEG compression, Gaussian blur, and state-of-the-art diffusion-based adversarial purification.
\end{itemize}
\section{Related Work}

\paragraph{Adversarial Attack}
Adversarial attacks introduce subtle perturbations to machine learning models, altering outputs while remaining imperceptible to humans~\cite{guo2024white}. Key methods include Fast Gradient Sign Method~\cite{goodfellow2014explaining}, Projected Gradient Descent~\cite{madry2017towards}, and Carlini-Wagner attacks~\cite{carlini2017towards}. These approaches have expanded beyond classification to generative models~\cite{guo2024grey,guo2024artwork}, including face recognition systems~\cite{10100731, 10382699, 10447890, xue2023diffusion, kang2023diffattack, dai2024advdiff, chen2023advdiffuser, 10716799, chen2023trojdiff}. Recent surveys highlight vulnerabilities in biometrics and autonomous systems~\cite{10.1145/3721479}, underscoring the need for our proposed protection mechanism.

\paragraph{Diffusion Model}
Diffusion models generate images through a forward noise-adding process and a reverse denoising process, avoiding GANs' mode collapse issues~\cite{ho2020denoising, song2020score, song2019generative, sohl2015deep}. Their high-fidelity outputs position them as state-of-the-art alternatives~\cite{dhariwal2021diffusion} with recent innovations enhancing realism~\cite{rombach2022high, podell2023sdxl, wang2023prolificdreamer, esser2024scaling}. For face swapping, models like DiffFace use conditional denoising to transfer identities while preserving target attributes~\cite{kim2025diffface}. While their photorealism is ideal for many applications, their robustness against attacks remains underexplored.

\paragraph{Face Swapping Methods Based on Deep Learning}
Face swapping has evolved from early facial landmark methods to GAN-based approaches like DeepFakes, which improved realism but often introduced perceptible artifacts~\cite{chen2020simswap, Zhu_2021_CVPR, li2020advancing,nirkin2019fsgan, 9103116, xu2022styleswap}. Diffusion models have enhanced quality through iterative denoising for seamless identity blending~\cite{kim2025diffface}, while recent advancements address pose variations through inpainting and CLIP-based feature disentanglement~\cite{han2024faceadapterpretraineddiffusion, baliah2024realistic, zhao2023diffswap}. This improved quality intensifies concerns regarding misuse, necessitating robust protective measures.

\paragraph{Adversarial Attack on Face Swapping Models}
Adversarial attacks on face swapping aim to disrupt identity transfer, aiding deepfake detection or privacy protection. Most research targets GAN-based models using gradient-based perturbations or substitute models~\cite{10100731, 10382699, 10447890}, with approaches like transferable attacks via substitute GANs~\cite{10100731}. Diffusion models' iterative denoising processes present unique challenges, highlighting a critical gap which our proposed method addresses by exploiting model-specific vulnerabilities.

\paragraph{Adversarial Attack on Diffusion Model}
As diffusion models gain popularity, tailored attacks are emerging~\cite{lei2025instant}. Methods like DiffAttack generate latent space perturbations by exploiting noise schedule sensitivities~\cite{10716799}, while other approaches such as Trojdiff~\cite{chen2023trojdiff}, Advdiffuser~\cite{chen2023advdiffuser}, Advdiff~\cite{dai2024advdiff}, Diff-PGD~\cite{xue2023diffusion}, and Kang's DiffAttack~\cite{kang2023diffattack} target specific tasks. Recent surveys emphasize societal risks of unsecured diffusion models~\cite{10.1145/3721479}. Our work pioneers attacks tailored to diffusion-based face swapping with a novel loss function.
\section{Preliminary}

\subsection{Diffusion model}
Denoising Diffusion Probabilistic Models (DDPM)~\cite{ho2020denoising} synthesize images by learning the reverse distribution of a Markov chain that incrementally adds Gaussian noise to the original image  \( \bm{x}_0\). The forward process is expressed as:
\begin{equation}\label{eq:forward}
    \bm{x}_t = \sqrt{\bar{\alpha}_t} \bm{x}_0 + \sqrt{1 - \bar{\alpha}_t} \bm{\epsilon}, \quad \bm{\epsilon} \sim \mathcal{N}(0, \mathbf{I}),
\end{equation}
where \(\bar{\alpha}_t\) governs the noise schedule. The reverse denoising step is formulated as:
\begin{equation}\label{eq:reverse}
    \hat{\bm{x}}_{t-1} = \frac{1}{\sqrt{\alpha_t}} \left( \bm{x}_t - \frac{\beta_t}{\sqrt{1 - \bar{\alpha}_t}} \bm{\epsilon}_{\bm{\theta}^*}(\bm{x}_t, t) \right) + \sqrt{\beta_t} \bm{\epsilon},
\end{equation}
with the model parameters \(\bm{\theta}^*\) optimized to minimize the mean squared error between predicted and actual noise:
\begin{equation}
    \bm{\theta}^* = \arg\min_\theta \mathbb{E}_{\bm{x}_0, t, \bm{\epsilon}} \left[ \|\bm{\epsilon} - \bm{\epsilon}_\theta(\bm{x}_t, t)\|_2^2 \right].
\end{equation}
To accelerate inference, Denoising Diffusion Implicit Models (DDIM)~\cite{songDDIM} employ a non-Markovian process, modifying the denoising step as:
\begin{equation}
    \hat{\bm{x}}_{t-1} = \sqrt{\bar{\alpha}_{t-1}} \left( \frac{\bm{x}_t - \sqrt{1 - \bar{\alpha}_t} \bm{\epsilon}_{\theta^*}(\bm{x}_t, t)}{\sqrt{\bar{\alpha}_t}} \right) + \sqrt{1 - \bar{\alpha}_{t-1} - \sigma_t^2} \bm{\epsilon}_{\theta^*}(\bm{x}_t, t) + \sigma_t \bm{\epsilon},
\end{equation}
where \(\sigma_t = \eta \cdot \frac{1 - \bar{\alpha}_{t-1}}{1 - \bar{\alpha}_t} \cdot \beta_t\). Setting \(\eta = 0\) yields a deterministic process, while \(\eta = 1\) aligns with DDPM’s stochastic behavior.

\subsection{Conditional Diffusion for Face Swapping}
Conditional diffusion models enhance standard diffusion processes by integrating auxiliary conditioning signals, such as identity embeddings, to steer image generation. In face swapping, the goal is to synthesize an image \(\hat{x}_0\) that combines the identity of a source face, encoded as a condition \(c\) (e.g., extracted via ArcFace~\cite{deng2019arcface}), with the structural attributes (e.g., pose, expression) of a target face. The noise prediction network is adapted to incorporate this conditioning, formulated as \(\epsilon_\theta(x_t, t, c)\), where \(x_t\) is the noisy image at timestep \(t\), and \(\theta\) denotes the model parameters. The training objective minimizes the mean squared error between predicted and actual noise:
\begin{equation}
    L = \mathbb{E}_{x_0, t, \epsilon, c} \left[ \| \epsilon - \epsilon_\theta(x_t, t, c) \|_2^2 \right]
\end{equation}
During inference, the denoising process iteratively refines \(x_t\) into \(\hat{x}_0\), guided by \(c\) to ensure identity fidelity. This approach leverages the generative power of diffusion models while enabling precise control over identity transfer, establishing it as a pivotal technique in diffusion-based face swapping frameworks~\cite{kim2025diffface,baliah2024realistic,han2024faceadapterpretraineddiffusion,zhao2023diffswap}.

\subsection{Classifier-Free Guidance}
Classifier-free guidance improves the quality of samples generated by conditional diffusion models without relying on an external classifier. The method trains a single model to jointly learn the conditional distribution \(p_\theta(x_t | c)\) and the unconditional distribution \(p_\theta(x_t)\). This is achieved by occasionally replacing the condition \(c\) with a null token during training, allowing the model to implicitly capture both distributions. At inference, the noise prediction is adjusted using a guidance scale \(w > 0\):
\begin{equation}\label{eq:cfg}
    \tilde{\epsilon}_\theta(x_t, t, c) = (1 + w) \epsilon_\theta(x_t, t, c) - w \epsilon_\theta(x_t, t,\varnothing )
\end{equation}
where \(\epsilon_\theta(x_t, t, c)\) is the conditional prediction, \(\epsilon_\theta(x_t, t)\) is the unconditional prediction, and \(\varnothing\) is the null token. The parameter \(w\) controls the trade-off between sample fidelity and diversity, with higher values emphasizing adherence to the condition \(c\). In face swapping, this technique enhances identity preservation by amplifying the influence of the identity embedding, achieving high-fidelity results efficiently~\cite{ho2022classifier}.

\subsection{SDEdit}
SDEdit~\cite{meng2022sdedit} is a technique for image synthesis and editing that leverages stochastic differential equations (SDEs) that extends the framework of diffusion models, which are known for their ability to generate high-quality images by modeling a gradual noise addition and removal process, by incorporating user-provided guidance to control the generation or modification of images. In inpainting task, SDEdit starts with an existing image, applies a forward process to add controlled noise, using \autoref{eq:forward}, and then uses the reverse SDE with guidance to edit the image, that is using \autoref{eq:reverse} by replacing \(\bm{\epsilon}_{\bm{\theta}}(\bm{x}_t, t)\) with \autoref{eq:cfg}.

\begin{align}
    \hat{\bm{x}}_{t-1} 
    &= \frac{1}{\sqrt{\alpha_t}} \left( \bm{x}_t - \frac{\beta_t}{\sqrt{1 - \bar{\alpha}_t}} \left((1 + w) \epsilon_\theta(x_t, t, c) - w \epsilon_\theta(x_t, t,\varnothing )\right) \right) + \sqrt{\beta_t} \bm{\epsilon}\label{eq:SDEdit}
\end{align}
\section{Method}
\begin{figure}[t!]
    \centering
    \includegraphics[width=\linewidth]{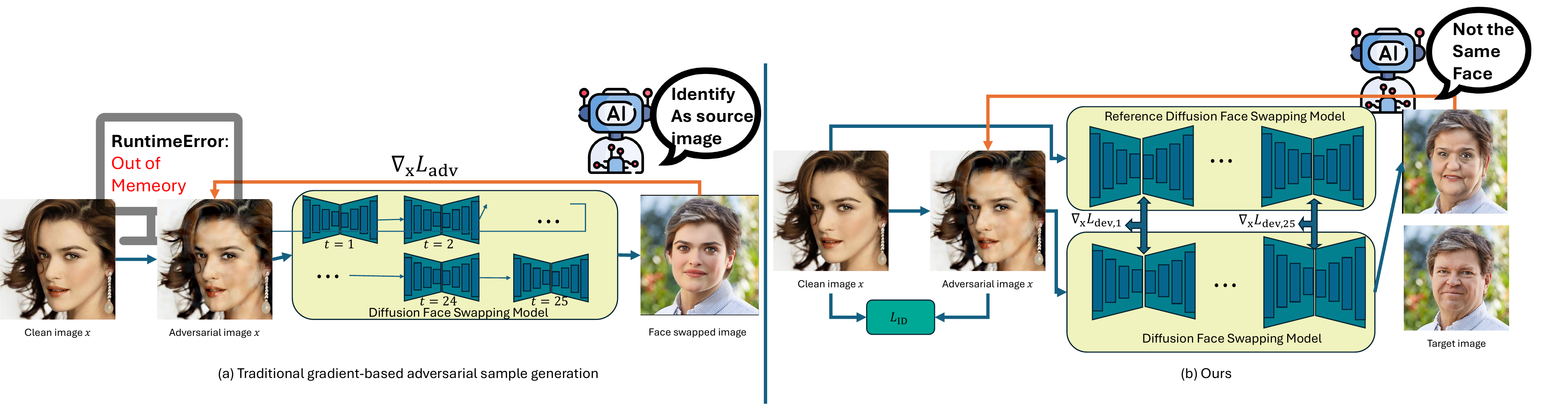}
    \caption{The pipeline of our method comparing to conventional gradient-based method}
    \label{fig:Pipeline}
    \vspace{-10pt}
\end{figure}

Developing effective protection against diffusion-based face swapping presents three fundamental challenges that define our approach. First, the multi-step iterative nature of diffusion models creates computational barriers—notably gradient instability and prohibitive memory requirements—that traditional adversarial methods cannot overcome. Second, the region-specific nature of facial manipulation in deepfake systems requires targeted perturbations rather than global patterns. Third, the diversity of diffusion-based implementations demands a protection mechanism that generalizes across architectures without requiring access to specific model weights.

\autoref{fig:Pipeline} illustrates our proposed framework, which strategically addresses these challenges through a two-pronged approach. Our method operates in the latent space of Latent Diffusion Models (LDMs), where we generate imperceptible perturbations that disrupt the identity transfer process while preserving image quality. The framework consists of two complementary components: (1) a face identity loss that targets the conditional mechanisms exploited in diffusion-based face manipulation, and (2) an inference-step averaging technique that efficiently generates robust perturbations without requiring prohibitive computational resources.

\subsection{Latent Space Perturbation}

The first challenge in protecting against diffusion-based face swapping is the prohibitive computational overhead typically associated with adversarial optimization. Our solution begins with a key insight: while pixel-space perturbations require high-dimensional optimization, latent space offers a compressed domain that dramatically reduces memory requirements while maintaining expressive power.

Our method leverages the latent space of Latent Diffusion Models (LDMs) to create protective perturbations with optimal efficiency. Given a source image $x$ that requires protection, we first transform it into its latent representation $z = \mathcal{E}(x)$ using the VAE encoder $\mathcal{E}$ of the LDM architecture. We then compute an adversarial perturbation $\delta$ through our optimization process with the associate loss function \(L\) and apply it to obtain the adversarial latent representation $z_{\text{adv}} = z + \delta$ with a budget $\epsilon$ in $p$ norm.
Finally, this perturbed latent is decoded back to image space using the VAE decoder $\mathcal{D}$, resulting in the protected image $x_{\text{adv}} = \mathcal{D}(z_{\text{adv}})$.
\begin{equation}
    \max_{\delta} L(\mathcal{D}(\mathcal{E}(x) +\delta)) \quad s.t. \quad\norm{\delta}_p \leq \epsilon
\end{equation}

This approach not only addresses the memory constraints but also provides a natural regularization effect. The structured nature of the latent space ensures that our perturbations, while sufficient to disrupt face swapping systems, remain imperceptible to human observers. This balance between effectiveness and invisibility is essential for real-world deployment of protection systems.

\subsection{Identity Loss (\texorpdfstring{$L_{\text{ID}}$}{L\_ID})}

The second challenge we address is the region-specific nature of facial manipulation. Our analysis of diffusion-based face swapping systems revealed that, unlike conventional image generation or editing, which use text prompt or style transfer, these models rely on specialized facial feature extractors to guide the identity transfer process. This insight led us to develop a targeted approach that specifically disrupts these conditioning pathways.

Diffusion-based face swapping systems, while built on diffusion models, operate differently from traditional models. Unlike conventional diffusion models that generate samples with limited control, face swapping systems integrate robust prior knowledge to guide the generation process. This information is processed through task-specific models—such as face recognition and facial keypoint detection—targeting distinct facial regions. These specialized extractors facilitate precise identity transfer from a source to a target image while minimizing visible artifacts, leveraging various techniques to manipulate identity features across facial areas.

To quantify the discrepancy in identity, we extract deep facial embeddings using state-of-the-art face recognition architectures that incorporate large-margin angular projections in their feature spaces . These embeddings effectively capture the discriminative characteristics of faces that the conditioning mechanisms of diffusion models employ during identity transfer. The identity loss is defined as:
\begin{equation}\label{eq:ID_Loss}
    L_{\text{ID}} (z_{\text{adv}},x) = 1 - \cos(f(\mathcal{D}(z_{\text{adv}})), f(x))
\end{equation}
where $f(\cdot)$ represents the embedding function that maps facial images to a normalized hypersphere, $\mathcal{D}(\cdot)$ denotes the decoding of the adversarial latent representation, and $\cos(\cdot,\cdot)$ measures the angular similarity between the resulting identity vectors. Maximizing this loss ensures that the identity signal derived from the perturbed image becomes misaligned with the conditioning pathways in the face-swapping process, thereby effectively inhibiting unauthorized identity transfers while preserving the overall visual fidelity of the image.
\vspace{-10pt}

\subsection{Average Deviate Loss (\texorpdfstring{$L_{\text{dev}}$}{L\_dev})}
\vspace{-10pt}

The third and most technically challenging obstacle we encountered was the gradient instability inherent in diffusion model optimization. Our initial experiments revealed that naively computing gradients through the entire diffusion process led to vanishing gradients and prohibitive memory consumption—rendering traditional adversarial approaches infeasible.

We approached this problem by first analyzing how diffusion models generate images.
Intuitively, each sample generated by a diffusion model follows a distinct trajectory that begins from a Gaussian distribution and terminates in the image space. Consequently, the trajectories associated with clean and adversarial images can be leveraged to construct adversarial attacks by perturbing the points along the adversarial trajectory toward those of the clean image. In doing so, it is ensured—according to~\autoref{eq:reverse}—that the final outputs \(z_{0,\text{adv}}\) and \(z_0\) differ. Specifically, for each timestep \(t\), we compute the deviation loss as
\begin{equation}\label{eq:Dev_Loss}
    L_{\text{dev}, t} (z_{\text{adv}},x)= \text{MSE}\left( \epsilon_\theta(z_t, t, c(\mathcal{D}(z_{\text{adv}}))), \epsilon_\theta(z_t, t, c(x)) \right),
\end{equation}
Nonetheless, several challenges arise from computing \(L_{\text{dev}, t}\) at every timestep. Diffusion models operate over a sequence of timesteps \(t = 1, 2, \ldots, T\), whereby an initial image \(z_0\) is progressively degraded into \(z_1, z_2, \ldots, z_T\) via the forward process (as specified in~\autoref{eq:forward}) and subsequently reconstructed in the reverse process (refer to~\autoref{eq:reverse}). Utilizing conventional gradient-based methods to differentiate any loss function \(L\) with respect to \(z_{\text{adv}}\) necessitates traversing a deep computational graph where each \(z_t\) depends on \(z_{t-1}\), thereby forming an extensive dependency chain across \(T\) timesteps. This extensive chain exacerbates two critical issues during backpropagation: (1) the vanishing gradient problem, and (2) high memory consumption due to the need to store intermediate states and gradients.

In computing the gradient of \(L_{\text{dev}, t}\) with respect to \(z_{\text{adv}}\), one must backpropagate through the entire diffusion process:
\begin{equation}
    \frac{\partial L_{\text{dev}, t}}{\partial z} = \frac{\partial L_{\text{dev}, t}}{\partial \epsilon_\theta(z_t, t, c^\prime_{\text{id}})} \cdot \frac{\partial \epsilon_\theta(z_t, t, c^\prime_{\text{id}})}{\partial z_t} \cdot \frac{\partial z_t}{\partial z_{\text{adv}}} + \text{(similar term for } c^\prime_{\text{id}}\text{)},
\end{equation}
with \(\frac{\partial z_t}{\partial z} = \sqrt{\bar{\alpha}_t}\). Since \(z_t\) inherently depends on all preceding timesteps—as is characteristic of Markovian processes—the dependency chain amplifies the problems of vanishing gradients and excessive memory demands.

To mitigate these issues, we detach \(z_t\) from the loss computation in~\autoref{eq:Dev_Loss} at every timestep, effectively treating \(z_t\) as a constant input to \(\epsilon_\theta\) (i.e., setting \(\frac{\partial z_t}{\partial z_{\text{adv}}} = 0\)). Consequently, the gradient \(\frac{\partial L_{\text{dev}, t}}{\partial z_{\text{adv}}}\) is computed solely through other paths, such as via \(c'_{\text{id}}\), which is a function of \(z_{\text{adv}}\). More specifically, the gradient with respect to \(z_{\text{adv}}\) is expressed as
\begin{equation}
    \nabla_{z_{\text{adv}}} L_{\text{dev}, t} = \frac{\partial L_{\text{dev}, t}}{\partial \epsilon_\theta(z_t, t, c^\prime_{\text{id}})} \cdot \frac{\partial \epsilon_\theta(z_t, t, c'_{\text{id}})}{\partial c'_{\text{id}}} \cdot \frac{\partial c'_{\text{id}}}{\partial x_{\text{adv}}} \cdot \frac{\partial x_{\text{adv}}}{\partial z_{\text{adv}}},
\end{equation}
where \(\frac{\partial x_{\text{adv}}}{\partial z_{\text{adv}}} = \frac{\partial \mathcal{D}(z_{\text{adv}})}{\partial z_{\text{adv}}}\). This gradient is then employed to update \(z_{\text{adv}}\), while the detachment of \(z_t\) prevents backpropagation through the entire diffusion process.

Moreover, the noise prediction model \(\epsilon_\theta(z_t, t, c)\) utilizes a single set of shared parameters \(\theta\) across all timesteps to improve resource efficiency, rather than maintaining distinct parameters \(\theta_t\) for each individual timestep. This design implies that the outputs at different timesteps are highly correlated due to the common parameter set. To counteract the variability introduced by the differing trajectories from \(z_t\) to \(z_0\) over \(T\) timesteps, we propose averaging the deviation loss across all timesteps. The average deviation loss is thus defined as
\begin{equation}\label{eq:Avg_dev_loss}
    L_{\text{dev}}(z_{\text{adv}},x) = \frac{1}{T} \sum_{t=1}^T L_{\text{dev}, t}(z_{\text{adv}},x),
\end{equation}
with \(T\) representing the total number of denoising steps. This aggregated loss ensures that the diffusion process yields outputs misaligned with the original identity, effectively steering the face-swapped image to diverge significantly from that generated using a clean identity input.

\begin{algorithm}[t]
    \small
    \caption{}\label{Algo1}
    \begin{algorithmic}[1]
        \Require Source image $x$ to be attacked, latent diffusion encoder $\mathcal{E}$, step size $\alpha$, attack iterations $N$, diffusion inference time $T$, perturbation budget $\varepsilon$, regularization weight $\lambda$, Face Adapter $F(\cdot,\cdot)$, cosine similarity $f_{\text{cos}}(\cdot,\cdot)$, noise prediction model $\tilde{\epsilon}_\theta(z_t, t, c)$, identity encoder $E_{\text{id}}(\cdot)$ as MLP $\circ f_{\text{id}}$
        
        \State $z \gets \mathcal{E}(x)$ \Comment{Encoding source image into latent space}
        \State $c_{id} \gets E_{\text{id}}(x)$ \Comment{Extracting identity encoding of source}
        \State Initialize zero tensor $\delta$ \Comment{Preparing perturbation accumulator}
        \For{$n = 0, 1, 2, \ldots, N$}
            \State $z' \gets \text{SDEdit}(z + \delta)$ \Comment{Applying SDEdit for perturbation}
            \State $x' \gets \mathcal{D}(z')$ \Comment{Decoding perturbed latent to image}
            \State $c'_{id} \gets E_{\text{id}}(x')$ \Comment{Extracting identity of perturbed image}

            \For{$t = 0, 1, 2, \ldots, T$}
                \State $ L_{\text{id}} \gets f_{\text{cos}}(f_{\text{id}}(x'), f_{\text{id}}(x))$ \Comment{Identity loss~\autoref{eq:ID_Loss}}

                \State $ L_{\text{dev},t} \gets \text{MSE}(\epsilon_\theta(z_t, t, c'_{id}), \epsilon_\theta(z_t, t, c_{id}))$  \Comment{Deviation losss~\autoref{eq:Dev_Loss}}
                
                \State $\delta \gets \delta + \nabla_{z'}( L_{\text{id}} + \lambda L_{\text{dev},t} )$ \Comment{Accumulating gradients}
            \EndFor
        
            \State $\delta \gets \text{Avg}(\delta)$ \Comment{Averaging perturbations~\autoref{eq:Avg_dev_loss}}
            \State $z \gets z + \alpha \cdot \text{sign}(\delta)$ \Comment{Applying perturbation}
        \EndFor
        
        \State $x_{\text{adv}}, \gets \mathcal{D}(z)$ \Comment{Decoding final adversarial image}
        \State $x_{\text{advDiff}} \gets \mathcal{D}(\text{SDEdit}(z))$ \Comment{Decoding diffusion-enhanced adversarial image}
        \State \Return $x_{\text{adv}}, x_{\text{advDiff}}$
    \end{algorithmic}
\end{algorithm}

\vspace{-6pt}
\subsection{Purification Resistant Attack}
\vspace{-4pt}
SDEdit (\autoref{eq:SDEdit}) is employed within the latent space to strengthen our protection mechanism against potential purification strategies. Specifically, following the application of the forward diffusion process (\autoref{eq:forward}) to the latent variable \(z_0\), the subsequent backward process in SDEdit (\autoref{eq:SDEdit}) introduces additional stochasticity, thereby generating a modified latent variable \(z_0^\prime \neq z_0\). This augmentation enhances the robustness of perturbations, ensuring resilience against variations in the latent space. Moreover, this approach significantly improves the quality of adversarial images by mitigating perceptual distortions. As illustrated in \autoref{fig:compare}, the application of SDEdit to \(x_{\text{adv}}\) results in the generation of \(x_{\text{adv\_diff}}\), which preserves high visual fidelity and details from the original source image while effectively safeguarding facial identity information against face-swapping techniques.
\vspace{-4pt}
\subsection{Objective Function}
\vspace{-4pt}
Combining our identity loss and average deviate loss, we formulate a comprehensive objective function that balances protection effectiveness with visual imperceptibility:
\begin{gather}
    \max_{\delta} L_{\text{total}}(z + \delta,x) = L_{\text{ID}}(z + \delta,x) + \lambda L_{\text{dev}}(z + \delta,x) \quad \text{ s.t. } \| \delta \|_{\infty} \leq\varepsilon~\label{eq:total_loss}\\
    z_{\text{adv}} = \mathcal{P}_{B_{\infty}(z,\varepsilon)} \left[ z_{\text{adv}}+\alpha\sign\nabla_{z_{\text{adv}}} L_{\text{total}} \right]
\end{gather}
where \( \lambda \) is a hyperparameter balancing the contributions of the two losses, typically set based on empirical validation, \( \varepsilon  \) is the budget that the attack allow under the selected infinity norm \( \|\cdot \|_{\infty}\).
\section{Experiment and Results}
\begin{figure}[t!]
    \centering
    \includegraphics[width=\linewidth]{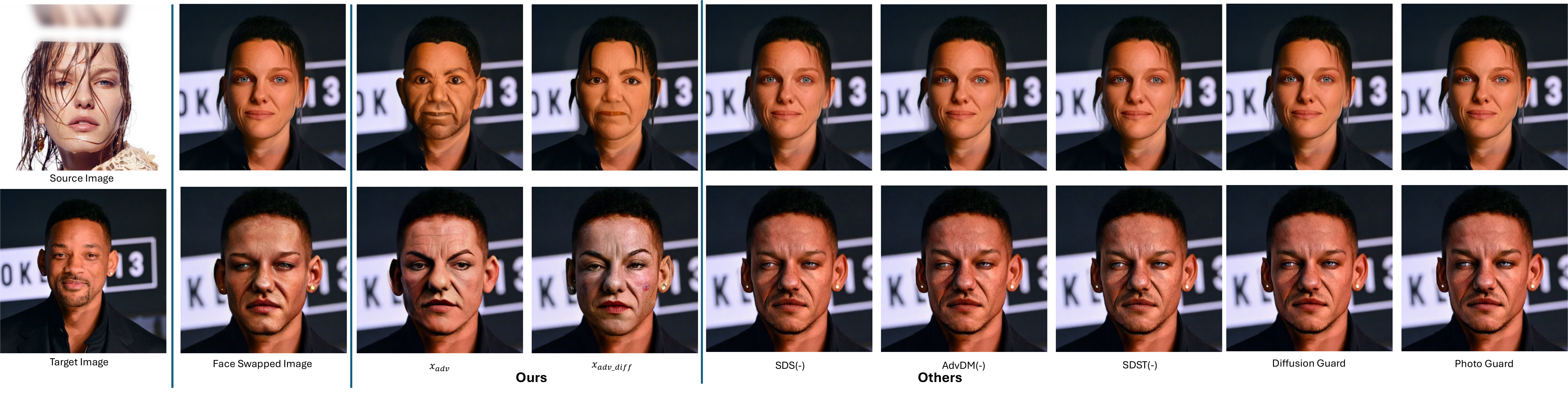}
    \caption{qualitative demonstration of our method comparing with SDS, AdvDM, SDST, DiffusionGuard, PhotoGuard across different diffusion-based face swapping techniques. The top row is face swapping result conducted on Face Adapter, whereas the bottom row is face swapping result conducted on REFace.}
    \label{fig:MainVis}
    \vspace{-16pt}
\end{figure}

\subsection{Experiment Setup}\label{subsec:ExpSetup}

We validate our method on 500 random CelebA-HQ~\cite{karras2017progressive} validation images for facial swapping, using StableDiffusion v1-5~\cite{Rombach_2022_CVPR} as pretrained LDM and ArcFace Glint360K~\cite{deng2019arcface} as face ID extractor. Images are resized to $3 \times 512 \times 512$ for StableDiffusion compatibility. We evaluate using LPIPS~\cite{zhang2018perceptual}, PSNR, SSIM~\cite{wang2004image}, and FID~\cite{heusel2017gans}. Facial identity similarity is measured via cosine similarity on ArcFace~\cite{deng2019arcface} embeddings, denoted as $CS_{SRC}$ and $CS_{ATT}$, where $CS_{SRC}$ refers to the similarity between swapped clean image and clean image, $CS_{ATT}$ refers to similarity between swapped attacked image and clean image. 
Higher LPIPS, FID, $CS_{ATT}$; and lower PSNR, SSIM, indicate better protection performance against face swapping. Specially, $CS_{SRC}$ serves as a baseline reference value for $CS_{ATT}$.
We benchmark against: PhotoGuard~\cite{salman2023raising} and DiffusionGuard~\cite{choi2024diffusionguardrobustdefensemalicious} (masked facial region perturbations); AdvDM~\cite{pmlr-v202-liang23g} (global image perturbations in denoising); Mist~\cite{liang2023mistimprovedadversarialexamples} and SDS~\cite{xue2023toward} (AdvDM refinements); and SDST~\cite{xue2023toward} (SDS variant incorporating target image considerations).
  
Our methods are implemented in PyTorch on a single Nvidia L40 GPU (48GB) with Ubuntu 22.04. We set hyperparameter $\lambda$ to $1.5 \times L_{\text{ID}} \times L_{\text{dev}}^{-1}$ to balance loss components. Attack iteration number $N=100$, diffusion inference time $T=25$, and SDEdit steps in \autoref{Algo1} is 3, following standard PGD attack, Face Adapter, and Diff-PGD settings. All baseline models use attack budget $\delta = 16/255$ and step size $\alpha = 1/255$, while our latent space method uses $\delta = 75/255$ and $\alpha = 10/255$ due to different scaling.

\subsection{Experiment Results}\label{subsec:ExpResult}

\begin{table}[!t]
\caption{Comparison of our method against several baselines. Arrows \( (\uparrow/ \downarrow)\) indicate whether higher or lower values represent better image quality for each metric. The \textbf{best result} is in bold and the \underline{second-best} result is underlined.}
\label{tab:main}
\resizebox{\linewidth}{!}{%
\begin{tabular}{l|cccccccccccccc}
\toprule
Metrics & AdvDM(-)~\cite{pmlr-v202-liang23g} & AdvDM(+)~\cite{pmlr-v202-liang23g} & DiffusionGuard~\cite{choi2024diffusionguardrobustdefensemalicious} & Mist~\cite{liang2023mistimprovedadversarialexamples} & PhotoGuard~\cite{salman2023raising} & SDS(-)~\cite{xue2023toward} & SDS(+)~\cite{xue2023toward} & SDST(-)~\cite{xue2023toward} & \(x_{\text{adv}}\) (Ours) & \(x_{\text{adv\_diff}}\) (Ours) \\
\midrule
LPIPS$\downarrow$      & 0.025    & 0.039   & 0.02    & 0.037          & 0.003       & 0.027  & 0.036    & 0.036    & \textbf{0.075}         & \underline{0.054} \\
PSNR$\uparrow$        & 30.914  & 29.531  & 35.319   & 29.441         & 41.13        & 30.596 & 29.39  & 29.67   & \textbf{24.58}        & \underline{26.296}          \\
SSIM$\uparrow$        & 0.97   & 0.957   & 0.987    & 0.957          & 0.996       & 0.967   & 0.957   & 0.958    & \textbf{0.91}           & \underline{0.931}           \\
FID$\downarrow$       & 6.665   & 8.091   & 1.434    & \textbf{16.116} & 0.253       & 7.837  & 7.392   & 13.365   & \underline{14.222}      & 5.590              \\
$\text{CS}_{SRC}$      & 0.119   & 0.119   & 0.119    & 0.124          & 0.119       & 0.119  & 0.119   & 0.119    & 0.119                & 0.119               \\
$\text{CS}_{ATT}$$\uparrow$  & 0.109   & 0.102    & 0.115    & 0.099          & 0.118       & 0.109  & 0.102   & 0.098    & \textbf{0.033}         & \underline{0.051}           \\
\bottomrule
\end{tabular}%
}
\vspace{-2em}
\end{table}

Table~\ref{tab:main} summarizes the performance of our two adversarial variants—\(x_{\text{adv}}\) and \(x_{\text{adv\_diff}}\)—against several baselines.
Existing methods such as AdvDM~\cite{pmlr-v202-liang23g}, DiffusionGuard~\cite{choi2024diffusionguardrobustdefensemalicious}, and PhotoGuard~\cite{salman2023raising} yield low LPIPS (e.g., PhotoGuard achieves 0.0025), high PSNR (up to 41.13), and near-perfect SSIM (0.9961), suggesting that the resulting images closely resemble those from clean inputs and thus preserve the underlying identity. In contrast, our methods intentionally perturb facial identity. The \(x_{\text{adv}}\) variant achieves a CosSim(Src2Attack) of 0.0327, markedly lower than baseline values (approximately 0.0976–0.1184). Meanwhile, \(x_{\text{adv\_diff}}\) offers a balanced trade-off with LPIPS, PSNR, SSIM, and FID values of 0.0537, 26.30, 0.9314, and 5.59, respectively, and a CosSim(Src2Attack) of 0.0512. Overall, our results validate the efficacy of our framework in significantly disrupting face swapping. These findings underscore the potential of our approach to safeguard facial identity by inducing substantial alterations that thwart unauthorized manipulations. These quantitative results demonstrate that our adversarial attack framework introduces decisive perceptual, structural, and distributional shifts, effectively impeding unauthorized face swapping operations.

\subsection{Transferability Attack Against REFace}
To assess the generalizability of our adversarial attack framework beyond our primary experimental setting, we evaluate its transferability when deployed against the REFace model~\cite{baliah2024realistic}. Table~\ref{tab:REFace} presents the quantitative results on REFace for both of our proposed variants, \(x_{\text{adv}}\) and \(x_{\text{adv\_diff}}\), alongside several state-of-the-art baselines on several metrics that are as same as Table~\ref{tab:main}.

Our \(x_{\text{adv}}\) variant demonstrates remarkable transferability on REFace. It achieves an LPIPS of 0.1034—the highest among all compared methods—together with the lowest PSNR (23.1685) and SSIM (0.8555), and an FID of 4.8815. Most notably, \(x_{\text{adv}}\) reduces the CosSim(Src2Attack) to 0.1972, a dramatic decrease compared to the baseline values, which lie in the range of 0.3401 to 0.3876. These results indicate that the facial identity embedded by the adversarial source is considerably disrupted, thereby impeding the reconstruction of the original identity during the face swapping process. Although the \(x_{\text{adv\_diff}}\) variant achieves competitive performance—with an LPIPS of 0.094, PSNR of 23.7357, SSIM of 0.8648, FID of 2.9137, and CosSim(Src2Attack) of 0.2416—the more aggressive \(x_{\text{adv}}\) variant offers the most pronounced identity disruption.

Overall, the transferability experiments on REFace corroborate the robustness of our adversarial framework. Despite slight variations in image quality metrics across methods, our approach consistently induces significant perceptual, structural, and distributional shifts necessary for reliable identity disruption. These findings underscore that our proposed attack can effectively generalize across different face swapping pipelines, thereby enhancing the overall resilience of facial information protection against malicious manipulation.

\begin{table}[t!]
\caption{Transferability on REFace~\cite{baliah2024realistic} Comparison of our method against several baselines. Arrows \( (\uparrow/ \downarrow)\) indicate whether higher or lower values represent better image quality for each image quality metric. The \textbf{best result} in each group is in bold, and the \underline{second-best} result is underlined.}
\label{tab:REFace}
\centering
\small
\setlength{\tabcolsep}{0.5mm}
\resizebox{\linewidth}{!}{%
\begin{tabular}{l|cccccccccc}
\toprule
Metrics & AdvDM(-)~\cite{pmlr-v202-liang23g} & AdvDM(+)~\cite{pmlr-v202-liang23g} & DiffusionGuard~\cite{choi2024diffusionguardrobustdefensemalicious} & Mist~\cite{liang2023mistimprovedadversarialexamples} & PhotoGuard~\cite{salman2023raising} & SDS(-)~\cite{xue2023toward} & SDS(+)~\cite{xue2023toward} & SDST(-)~\cite{xue2023toward} & \(x_{\text{adv}}\) (Ours) & \(x_{\text{adv\_diff}}\) (Ours) \\
\midrule
LPIPS$\downarrow$      & 0.0785        & 0.0828        & 0.0766     & 0.0865         & 0.0756        & 0.0791   & 0.082   & 0.0851    & \textbf{0.1034}       & \underline{0.094}        \\
PSNR$\uparrow$        & 25.2026   & 24.8509   & 25.3325     & 24.5378       & 25.4215     & 25.1574   & 24.907       & 24.6652   & \textbf{23.1685}           & \underline{23.7357}               \\
SSIM$\uparrow$       & 0.8787   & 0.874   & 0.8806     & 0.8707         & 0.8815   & 0.878  & 0.875   & 0.8721       & \textbf{0.8555}           & \underline{0.8648}            \\
FID$\downarrow$       & 2.8027   & 3.4879   & 2.3851    & \underline{4.254}        & 2.2954             & 2.9379  & 3.1952       & 3.9487      & \textbf{4.8815 }      & 2.9137               \\
$\text{CS}_{SRC}$       & 0.3886   & 0.3885   & 0.3886     & 0.389          & 0.3884         &  0.3886    & 0.3885    &0.3886     & 0.3887             & 0.3886             \\
$\text{CS}_{ATT}$$\uparrow$  & 0.38   & 0.3674   & 0.3859     & 0.3401         & 0.3876        & 0.3792  & 0.3684     & 0.346    &  \textbf{0.1972}         & \underline{0.2416}             \\
\bottomrule
\end{tabular}%
}
\vspace{-10pt}
\end{table}

\begin{table}[t!]
\caption{Left column: The ablation studies of the \(L_\text{ID}\) and \(L_\text{dev}\). Middle and right column: Our proposed method against typical defense methods: Gaussian Blur and Jpeg Compression}
\label{tab:Defense}
\centering
\tiny
\setlength{\tabcolsep}{0.5mm}
\resizebox{\linewidth}{!}{%
\begin{tabular}{l|cccc|ccc|ccc|cc}
\toprule
    & \multicolumn{4}{c|}{Ablation} & \multicolumn{3}{c|}{Gaussian Blur} & \multicolumn{3}{c|}{JPEG Compression} & \multicolumn{2}{c|}{No Defense}\\
\midrule
Metrics & \(L_{\text{ID}}\ x_{\text{adv}}\) & \(L_{\text{ID}}\ x_{\text{adv\_diff}}\) & \(L_{\text{dev}}\ x_{\text{adv}}\) & \(L_{\text{dev}}\ x_{\text{adv\_diff}}\) 
& \(x\) & \(x_{\text{adv}}\) & \(x_{\text{adv\_diff}}\) 
& \(x\) & \(x_{\text{adv}}\) & \(x_{\text{adv\_diff}}\) &  \(x_{\text{adv}}\) & \(x_{\text{adv\_diff}}\)\\
\midrule
LPIPS$\downarrow$            & 0.074  & 0.054  & 0.109  & 0.074  & 0.012  & 0.075  & 0.056  & 0.0002  & 0.075  & 0.054 & 0.075         & 0.054\\
PSNR$\uparrow$               & 24.598 & 26.238 & 22.818 & 24.682 & 34.836 & 24.53  & 26.132 & 52.731  & 24.567 & 26.292 & 24.58        & 26.296\\
SSIM$\uparrow$               & 0.911  & 0.931  & 0.888  & 0.918  & 0.985  & 0.912  & 0.931  & 1       & 0.91   & 0.931 & 0.91           & 0.931\\
FID$\downarrow$              & 13.38  & 5.394  & 51.031 & 20.813 & 1.695  & 15.18  & 6.346  & 0.019   & 14.243 & 5.611 & 14.222      & 5.590\\
$\text{CS}_{SRC}$            & 0.119  & 0.119  & 0.119  & 0.119  & 0.119  & 0.119  & 0.119  & 0.119   & 0.119  & 0.119 & 0.119    & 0.119 \\
$\text{CS}_{ATT}$$\uparrow$  & 0.033  & 0.051  & 0.067  & 0.084  & 0.123  & 0.034  & 0.052  & 0.119   & 0.033  & 0.051 & 0.033         & 0.051\\
\bottomrule
\end{tabular}%
}
\vspace{-19pt}
\end{table}

\section{Conclusion}
In this work, we introduced a robust, state-of-the-art adversarial defense framework tailored specifically to counteract diffusion-based face swapping attacks. Our method harnesses a dual-loss strategy combining identity loss and average deviate loss, which creates imperceptible yet highly effective perturbations in the latent space. By leveraging the conditional mechanisms inherent in diffusion models, our approach disrupts unauthorized facial identity transfer while preserving the overall visual fidelity of the source images. Extensive experimental evaluations on benchmark datasets demonstrate significant perceptual, structural, and distributional shifts between clean and manipulated images, confirming the method’s superior performance over existing defenses. Furthermore, our approach adapts seamlessly across diverse diffusion architectures, ensuring it works effectively in varied adversarial scenarios. Overall, our research establishes a practical and powerful countermeasure against deepfake technologies, advancing the field of proactive facial information protection against malicious manipulation. These promising results underscore the necessity of developing specialized defenses in response to ever-evolving deepfake techniques, thereby ensuring enhanced privacy and protection in real-world applications.
\newpage
\bibliographystyle{unsrt}
{\footnotesize\bibliography{main}}
\appendix

\section{Appendix}
\subsection{Limitations}
While the proposed method offers a promising defense against diffusion-based deepfake face swapping, it is not without limitations. The approach relies on strong assumptions, such as the effectiveness of latent space perturbations and the sufficiency of the dual-loss strategy, which may not hold universally across all diffusion models or image conditions. For instance, its robustness could falter against future or differently architected models, and its performance might degrade with low-quality images or those featuring occlusions. Additionally, the method's computational demands, though mitigated, could still hinder large-scale or real-time applications. From a societal perspective, while it enhances privacy by preventing unauthorized face swaps, it does not address broader privacy issues like data collection or surveillance, and its fairness across demographic groups remains unexamined, potentially perpetuating biases in protection efficacy. These limitations underscore the need for further research to broaden the method's applicability and ensure equitable protection.
\subsection{Broader impact}
The method introduced in this work, which employs adversarial attacks to shield facial images from diffusion-based deepfake face swapping, brings forth profound positive societal impacts that bolster privacy, security, and ethical technology development. Primarily, it enhances privacy protection by proactively defending against unauthorized face swapping, a critical advancement that prevents the emotional distress and reputational harm often inflicted by non-consensual deepfake content, thereby offering individuals a robust safeguard in an increasingly digital world. Moreover, by raising the difficulty of crafting convincing deepfakes, the approach significantly mitigates misinformation, curbing the potential for fabricated media to sway public opinion, or destabilize social trust, which is especially vital in an era where misinformation proliferates rapidly online. This method also empowers individuals, particularly vulnerable populations such as activists, journalists, and public figures, by granting them greater control over their digital identities, reducing the risks associated with malicious exploitation and fostering a safer, more equitable online environment. Additionally, it contributes to the advancement of ethical AI research by prioritizing user security and privacy, setting a commendable standard for responsible innovation that addresses the dual-use dilemmas posed by generative technologies. On the downside, the technique might limit legitimate uses of face swapping in entertainment or virtual reality and could spur an adversarial arms race as attackers devise new bypass methods.
\clearpage
\subsection{More Visual Illustrations}
\paragraph*{Comparison of ours and the existing across various face swapping models and target images} \ 

\begin{figure}[H]
    \centering
    \includegraphics[height=20cm,keepaspectratio]{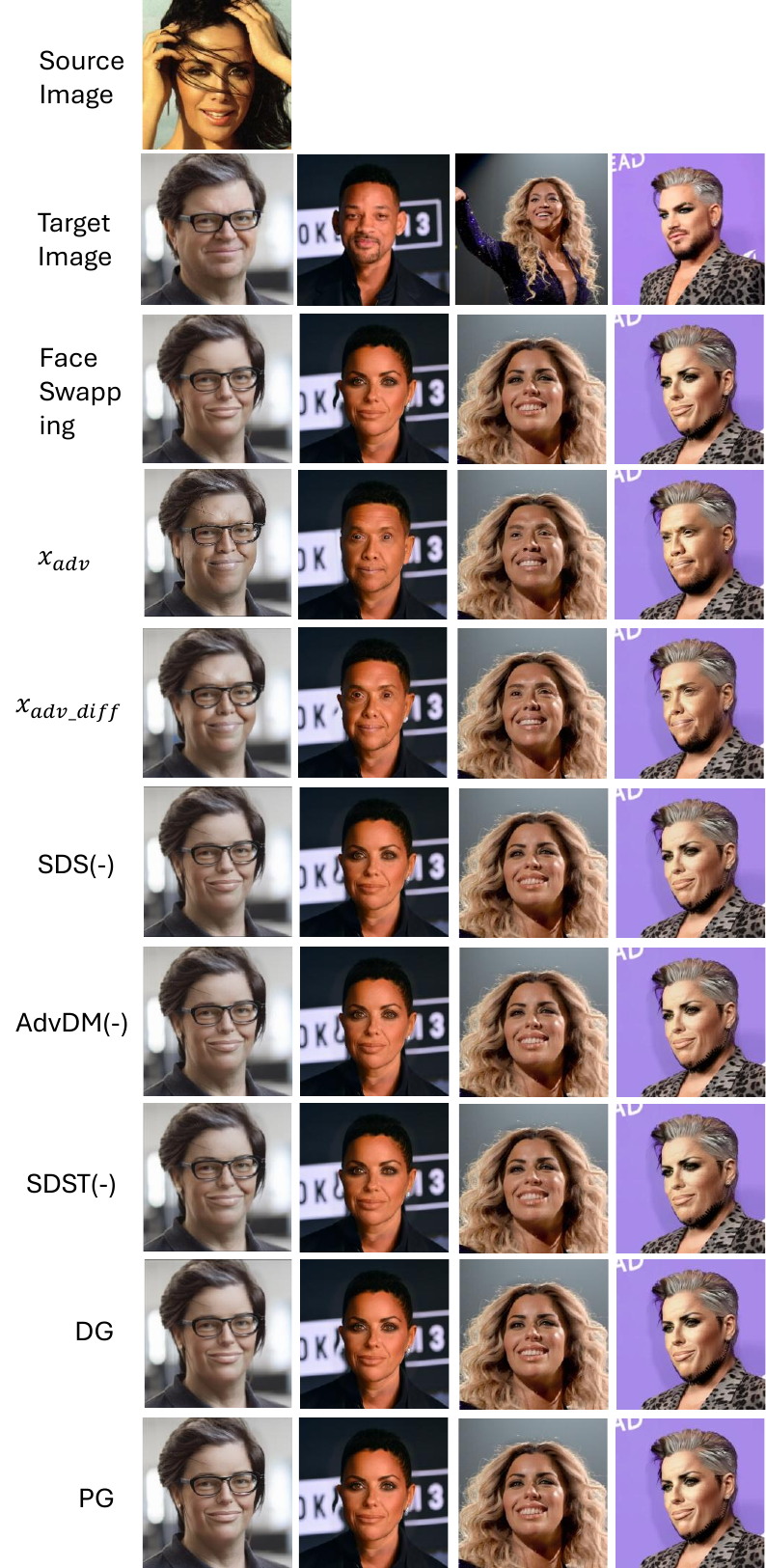}
    \caption{A qualitative visualization of our method comparing with existing methods on Face-Adapter. It is worth noting that, DG refers to DiffusionGuard, while PG refers to PhotoGuard.}
    \label{fig:app_faceAdapter}
\end{figure}
\newpage
\begin{figure}[H]
    \centering
    \includegraphics[height=20cm,keepaspectratio]{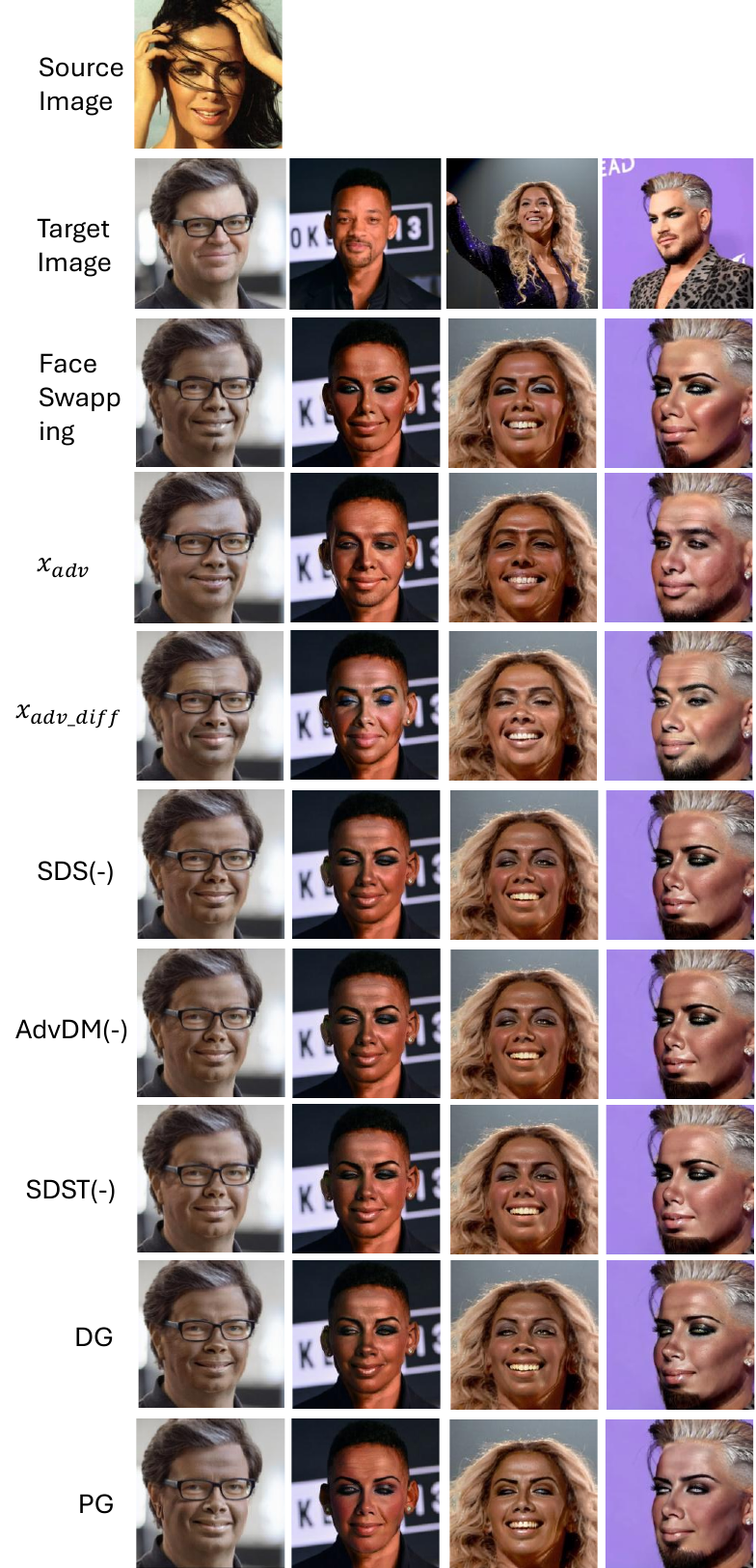}
    \caption{A qualitative visualization of our method comparing with existing methods on REFace. It is worth noting that, DG refers to DiffusionGuard, while PG refers to PhotoGuard.}
    \label{fig:app_REFace}
\end{figure}
\newpage
\begin{figure}[H]
    \centering
    \includegraphics[height=20cm,keepaspectratio]{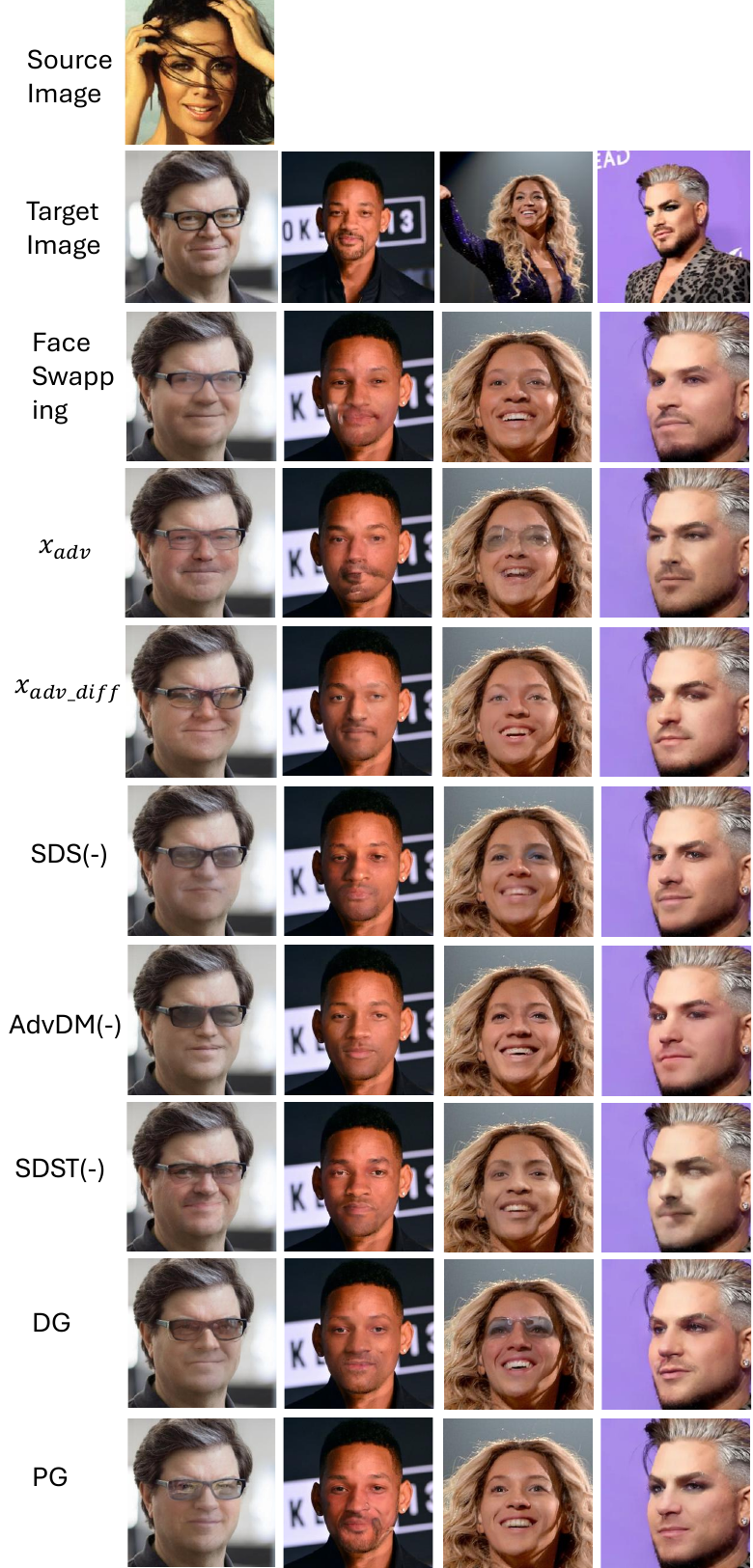}
    \caption{A qualitative visualization of our method comparing with existing methods on DiffSwap. It is worth noting that, DG refers to DiffusionGuard, while PG refers to PhotoGuard.}
    \label{fig:app_DiffSwap}
\end{figure}
\newpage
\paragraph{Ablation}\ 

\vspace{4em}
\begin{figure}[H]
    \centering
    \includegraphics[width=\linewidth]{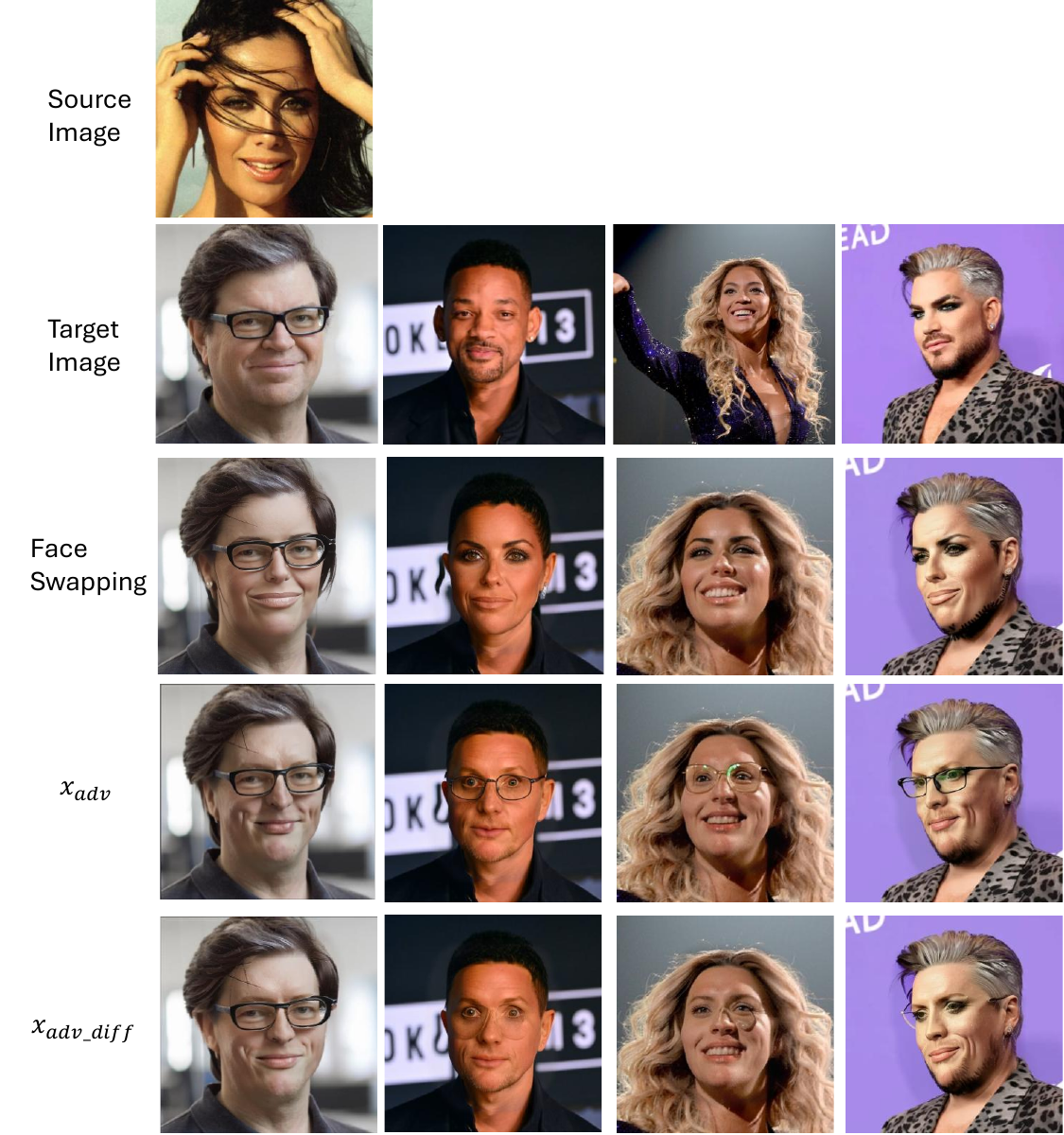}
    \caption{A visualization of ablation study on identity loss~($L_{\text{ID}}$). It is worth noting that, Face Swapping refers to directly face swap clean source image to target images with its corresponding columns. \(x_{adv}\) refers to the face swapped image using attacked image that crafted by identity loss only as source image. $x_{adv\_diff}$ refers to the face swapped image using attacked image that crafted by identity loss only and undergoes SDEdit as source image.}
    \label{fig:app_idLoss}
\end{figure}
\newpage

\begin{figure}[H]
    \centering
    \includegraphics[width=\linewidth]{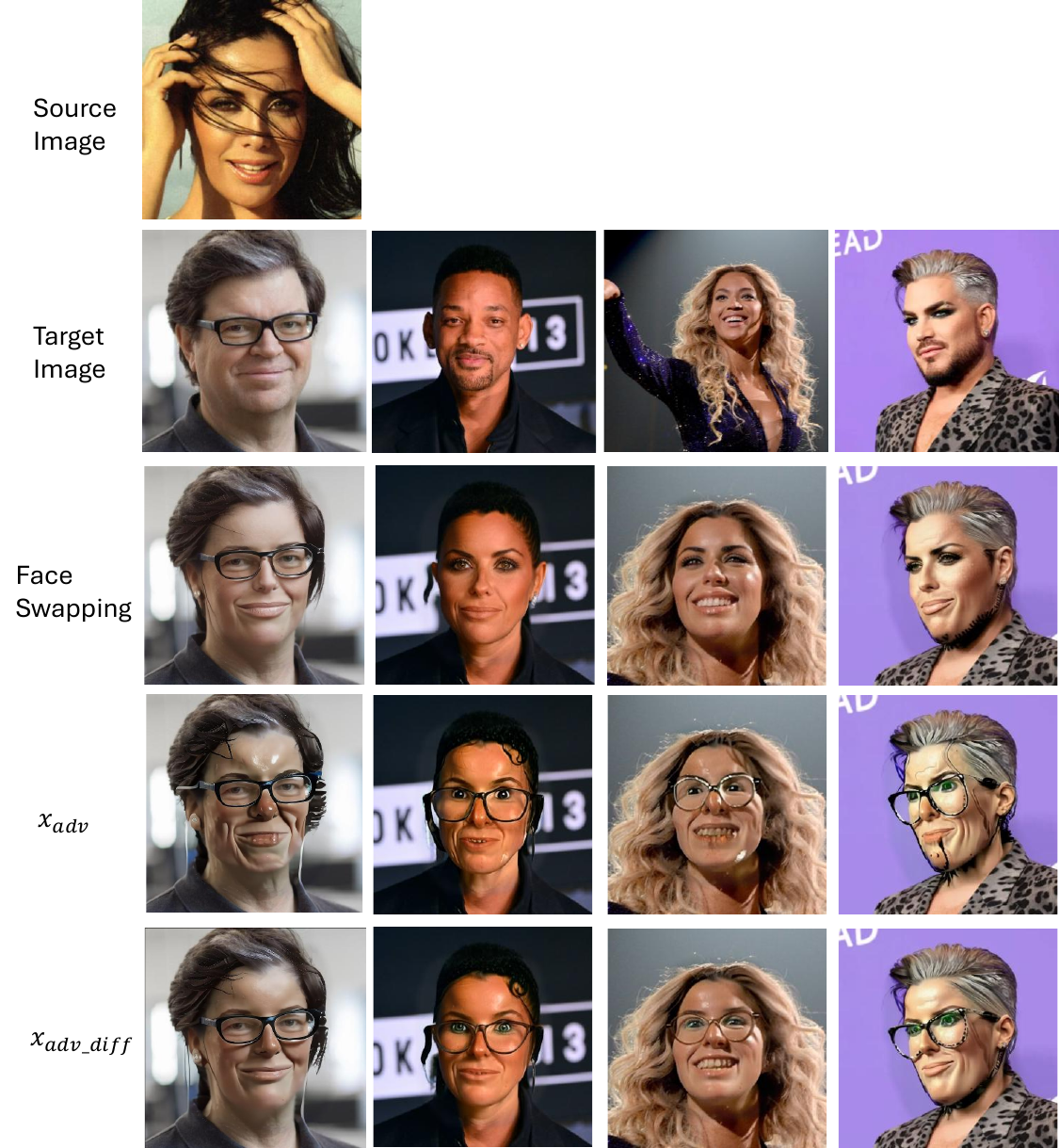}
    \caption{A visualization of ablation study on average deviate loss~($L_{\text{dev}}$). It is worth noting that, Face Swapping refers to directly face swap clean source image to target images with its corresponding columns. \(x_{adv}\) refers to the face swapped image using attacked image that crafted by average deviate loss only as source image. $x_{adv\_diff}$ refers to the face swapped image using attacked image that crafted by average deviate loss only and undergoes SDEdit as source image.}
    \label{fig:app_aveDevLoss}
\end{figure}
\newpage

\paragraph{Against defense}\ 
\vspace{1em}
\begin{figure}[H]
    \centering
    \includegraphics[width=\linewidth]{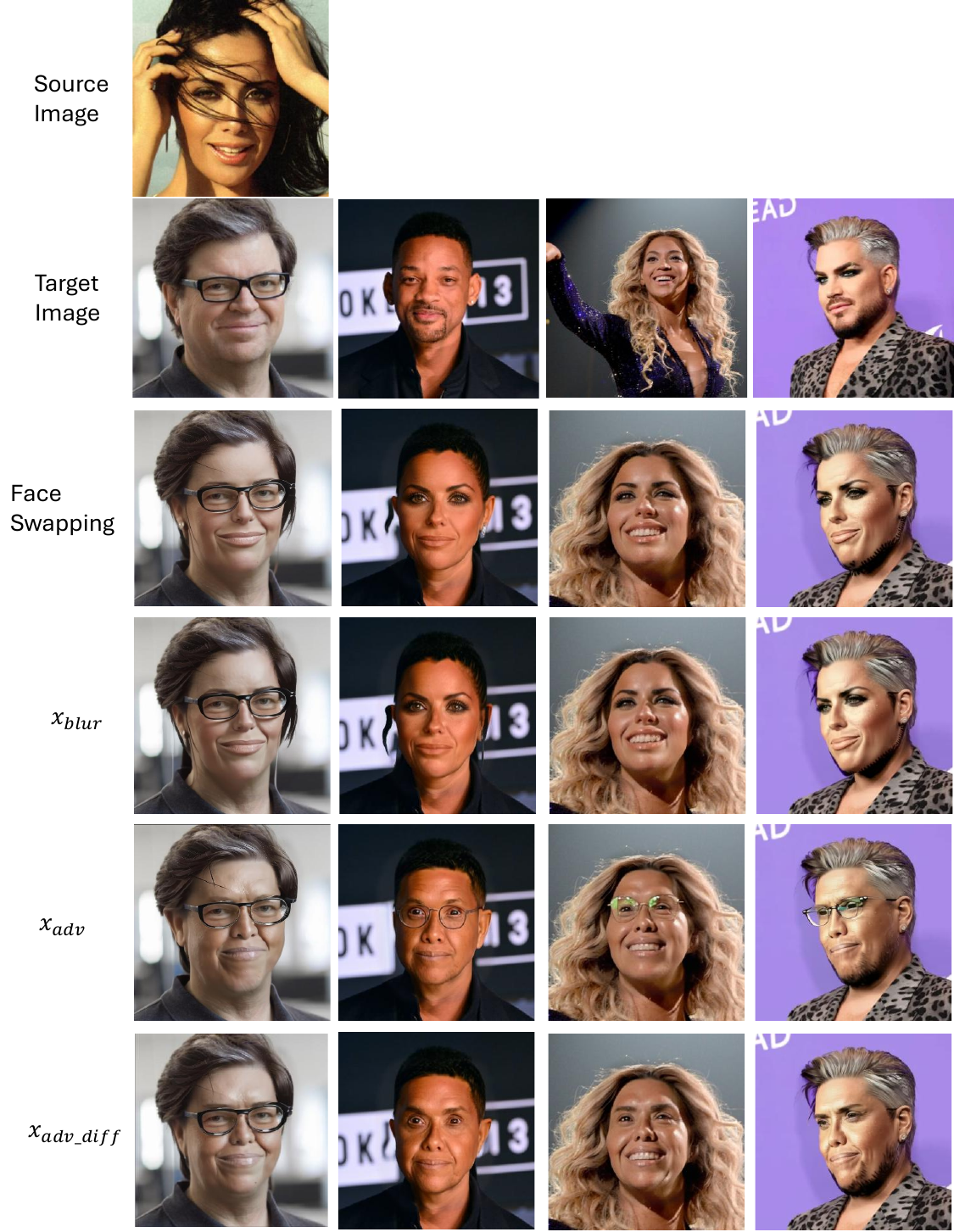}
    \caption{A visualization of our method against Gaussian Blur defense. It is worth noting that, Face Swapping refers to directly face swap clean source image to target images with its corresponding columns. $x_{blur}$ refers to the face swapped image using clean image that undergoes Gaussian Blur as source image. \(x_{adv}\) refers to the face swapped image using attacked image that crafted by our proposed method and undergoes Gaussian Blur as source image. $x_{adv\_diff}$ refers to the face swapped image using attacked image that crafted by our proposed method and undergoes SDEdit and Gaussian Blur as source image.}
    \label{fig:app_GaussianBlur}
\end{figure}
\newpage
\begin{figure}[H]
    \centering
    \includegraphics[width=\linewidth]{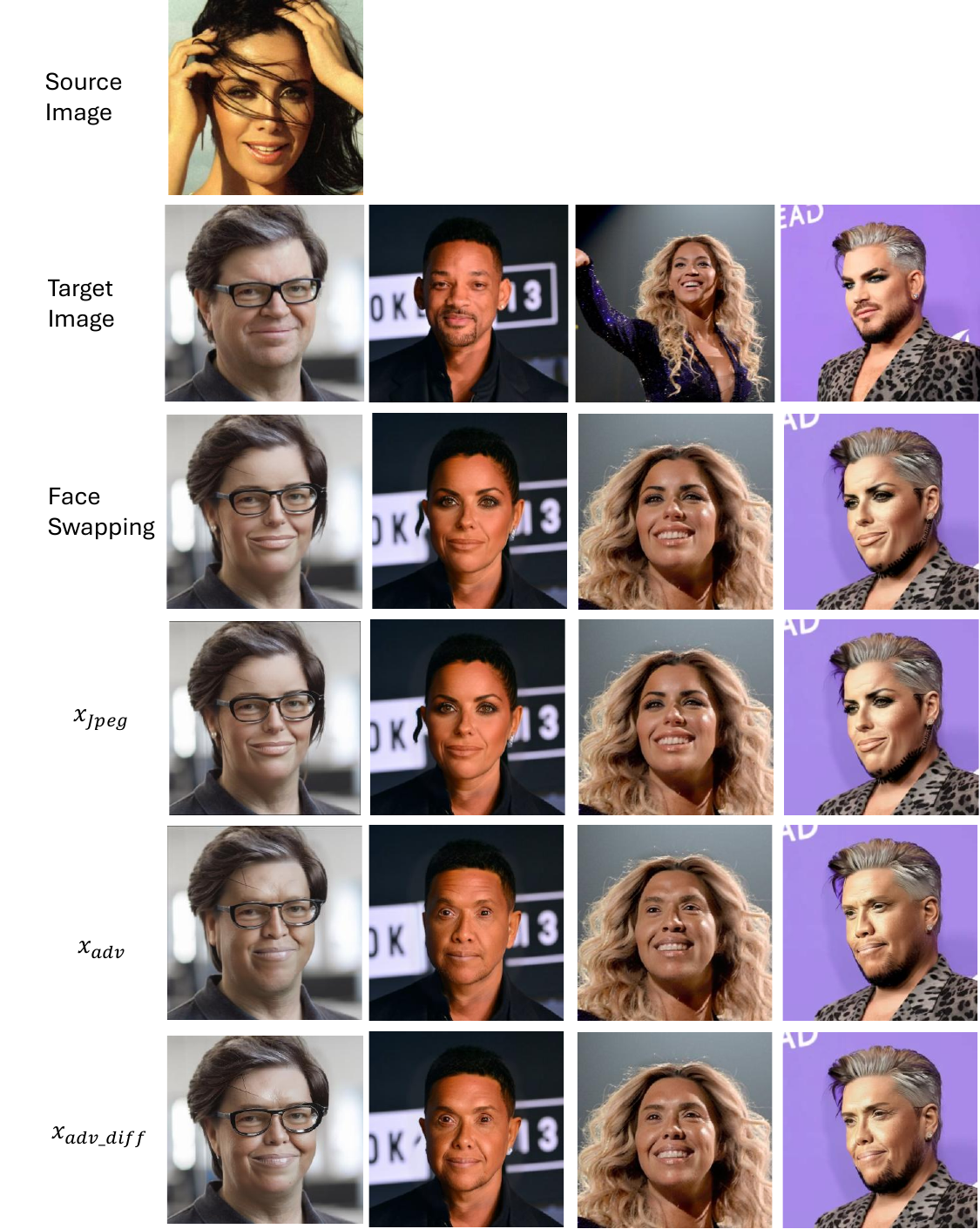}
    \caption{A visualization of our method against Jpeg Compression defense. It is worth noting that, Face Swapping refers to directly face swap clean source image to target images with its corresponding columns. $x_{Jpeg}$ refers to the face swapped image using clean image that undergoes Jpeg Compression as source image. 
    \(x_{adv}\) refers to the face swapped image using attacked image that crafted by our proposed method and undergoes Jpeg Compression as source image. $x_{adv\_diff}$ refers to the face swapped image using attacked image that crafted by our proposed method and undergoes SDEdit and Jpeg Compression as source image.}
    \label{fig:app_Jpeg}
\end{figure}
\end{document}